\pdfoutput=1

\documentclass[11pt]{article}
\usepackage{booktabs, makecell, tabularx}
\usepackage{makecell}    
\usepackage{tabularx}    
\newcolumntype{Y}{>{\centering\arraybackslash}X}  
\usepackage[final]{acl}
\usepackage{needspace} 
\usepackage{dblfloatfix} 
\usepackage{adjustbox}   
\usepackage{cuted}
\usepackage{caption}  

\usepackage{amsmath}   
\usepackage{pifont}
\usepackage{amsmath} 
\usepackage{booktabs} 
\usepackage{cuted}    
\usepackage{capt-of}  
\usepackage{tabularx, makecell} 
\usepackage{booktabs} 
\usepackage{adjustbox} 
\usepackage{times}
\usepackage{latexsym}
\usepackage{subcaption}
\usepackage{colortbl}
\usepackage{booktabs}
\usepackage{array}
\usepackage{multirow}
\usepackage{enumerate}
\usepackage{adjustbox} 

\usepackage{xcolor}

\usepackage[T1]{fontenc}

\usepackage[utf8]{inputenc}

\usepackage{microtype}

\usepackage{inconsolata}
\usepackage{listings}
\usepackage{color}
\usepackage{algorithm} 
\usepackage{algpseudocode} 
\usepackage{graphicx} 
\usepackage{float}    
\usepackage{pifont}

\usepackage{times}
\usepackage{microtype}
\usepackage{epsfig}
\usepackage{float}
\usepackage{placeins}
\usepackage{color, colortbl}
\usepackage{enumitem}
\usepackage{tabularx}
\usepackage{xstring}
\usepackage{multirow}
\usepackage{xspace}
\usepackage{hyperref}
\usepackage{url}
\usepackage{xcolor}
\usepackage[hang,flushmargin]{footmisc}

\definecolor{codegreen}{rgb}{0,0.6,0}
\definecolor{codegray}{rgb}{0.5,0.5,0.5}
\definecolor{codepurple}{rgb}{0.58,0,0.82}
\definecolor{backcolour}{rgb}{0.95,0.95,0.92}
\definecolor{mypurple}{RGB}{200,192,248}
\definecolor{mypurpledeep}{RGB}{142,126,240}
\definecolor{mygreen}{RGB}{117,170,156}
\definecolor{myyellow}{RGB}{255,192,0}
\definecolor{myblue}{RGB}{57,143,255}
\definecolor{mygrey}{RGB}{231,230,230}
\definecolor{codey}{RGB}{220,220,170}
\definecolor{coder}{RGB}{206,145,120}
\definecolor{codeb}{RGB}{156,220,254}
\definecolor{codenum}{RGB}{204,204,204}
\definecolor{headerblue}{RGB}{224,235,245}

\usepackage{multicol} 
\usepackage{rotating}    

\usepackage[most]{tcolorbox} 

\newtcolorbox{analysisbox}[1][]{
    enhanced jigsaw,
    colback=white,
    colframe=blue!75!black,
    fonttitle=\bfseries,
    boxsep=5pt,
    left=5pt,
    right=5pt,
    top=5pt,
    bottom=5pt,
    title=#1,
}

\newtcolorbox{analysisboxcode}[1][]{
    enhanced jigsaw,
    colback=white,
    colframe=yellow!75!black,
    fonttitle=\bfseries,
    boxsep=5pt,
    left=5pt,
    right=5pt,
    top=5pt,
    bottom=5pt,
    title=#1,
}

\newcommand{\cmark}{\textcolor{mygreen}{\ding{51}}} 
\newcommand{\xmark}{\textcolor{mypurpledeep}{\ding{55}}} 

\lstdefinestyle{mystyle}{
    backgroundcolor=\color{backcolour},   
    commentstyle=\color{codegreen},
    keywordstyle=\color{magenta},
    numberstyle=\tiny\color{codegray},
    stringstyle=\color{codepurple},
    basicstyle=\footnotesize,
    breakatwhitespace=false,         
    breaklines=true,                 
    captionpos=b,                    
    keepspaces=true,                 
    numbers=left,                    
    numbersep=5pt,                  
    showspaces=false,                
    showstringspaces=false,
    showtabs=false,                  
    tabsize=2
}

\usepackage[most]{tcolorbox}
\usepackage{float}
\usepackage{xspace}
\tcbset{
  aibox/.style={
    width=474.18663pt,
    top=10pt,
    colback=white,
    colframe=black,
    colbacktitle=black,
    enhanced,
    center,
    attach boxed title to top left={yshift=-0.1in,xshift=0.15in},
    boxed title style={boxrule=0pt,colframe=white,},
  }
}
\newtcolorbox{AIbox}[2][]{aibox,title=#2,#1}

\def\github{\raisebox{-2pt}{\includegraphics[height=1.05em]{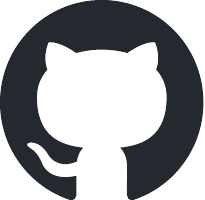}}}
\def\huggingface{\raisebox{-2pt}{\includegraphics[height=1.05em]{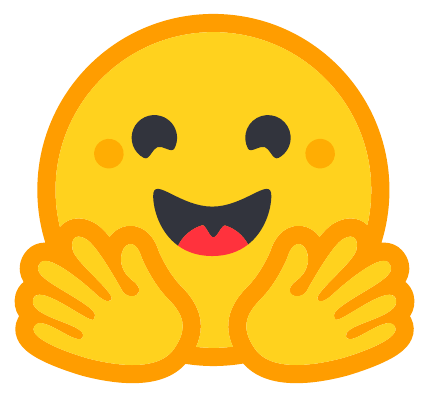}}}

\newcommand{\ghlink}{https://github.com/jbdel/RadEval}
\newcommand{\hflink}{https://huggingface.co/IAMJB/RadEvalModernBERT}

\title{RadEval: A framework for radiology text evaluation}

\author{
  Justin Xu$^{\spadesuit,}$\thanks{Equal contributions}
  \\\And
  Xi Zhang$^{\diamondsuit}$ 
  \\\And
  Javid Abderezaei$^{\heartsuit}$ 
  \\\And
  Julie Bauml$^{\heartsuit}$ 
  \\\AND
  Roger Boodoo$^{\heartsuit}$ 
      \\\And
    Fatemeh Haghighi$^{\heartsuit}$ 
    \\\And
Ali Ganjizadeh$^{\heartsuit}$ 
  \\\AND
  Eric Brattain$^{\heartsuit}$ 
      \\\And
  Dave Van Veen$^{\heartsuit}$ 
      \\\And
Zaiqiao Meng$^{\diamondsuit}$ 
      \\\AND
David Eyre$^{\spadesuit}$ 
\\\And  
  Jean-Benoit Delbrouck$^{\heartsuit,}$\footnotemark[1]
  \\\AND
  $^{\spadesuit}$University of Oxford \hspace{0.2cm}  $^{\diamondsuit}$University of Glasgow \hspace{0.2cm} 
 $^{\heartsuit}$HOPPR  \\
  \texttt{justin.xu@ndm.ox.ac.uk, jeanbenoit.delbrouck@hoppr.ai}\\
  \makebox[0pt][c]{%
    \begin{tabular}{c}
        \github\hspace{0.1em}      \url{\ghlink}\\
        \huggingface\hspace{0.1em} \url{\hflink}
    \end{tabular}
    }
}

\begin{document}
\maketitle

\begin{abstract}
We introduce RadEval, a unified, open-source framework for evaluating radiology texts. RadEval consolidates a diverse range of metrics -- from classic n‑gram overlap (BLEU, ROUGE) and contextual measures (BERTScore) to clinical concept-based scores (F1CheXbert, F1RadGraph, RaTEScore, SRR-BERT, TemporalEntityF1) and advanced LLM‑based evaluators (GREEN). We refine and standardize implementations, extend GREEN to support multiple imaging modalities with a more lightweight model, and pretrain a domain-specific radiology encoder -- demonstrating strong zero-shot retrieval performance. We also release a richly annotated expert dataset with over 450 clinically significant error labels and show how different metrics correlate with radiologist judgment. Finally, RadEval provides statistical testing tools and baseline model evaluations across multiple publicly available datasets, facilitating reproducibility and robust benchmarking in radiology report generation.
\end{abstract}

\section{Introduction}
\label{sec:intro}
Evaluating automated radiology report generation (RRG) systems remains a fundamental challenge in the development of safe, accurate, and clinically useful medical AI. Unlike general-purpose text generation tasks, RRG demands evaluation methods that can assess not only linguistic fluency but also clinical factuality, domain-specific terminology, uncertainty calibration, and diagnostic relevance. In recent years, the evaluation of radiology report generation has steadily progressed: initial studies relied on classic natural language generation (NLG) metrics such as BLEU and ROUGE~\cite{zhang2020optimizing, chen2020generating}; subsequent work emphasized clinical accuracy through disease-classification and natural language inference (NLI)-based metrics~\cite{miura2021improving}; this was followed by expert-annotated semantic graphs capturing entities and their relations~\cite{delbrouck2022improving}; and, most recently, by evaluation approaches that leverage large language models (LLM) \cite{ostmeier-etal-2024-green,bannur2024maira,huang2024fineradscore}. \\

\noindent Despite efforts to establish fair benchmarking, such as shared metric codebases released for challenge tracks~\cite{abacha2021overview, delbrouck2023overview, xu2024overview} and a public leaderboard~\cite{zhang2024rexrank}, there is still no open-source repository that reproduces the different factuality-focused metrics, whose scores can vary with implementation choices. For instance, earlier studies have computed BERTScore with different pretrained models and settings -- varying the number of layers or whether scores are rescaled with a baseline~\cite{zhang2019bertscore} -- or swapped in F1CheXbert embeddings for the calculation~\cite{smit2020combining}. Variants of the F1RadGraph metric likewise diverge depending on how they judge the correctness of entities and relations~\cite{delbrouck-etal-2022-improving}. Composite scores such as RadCliQ~\cite{yu2023evaluating} are similarly challenging to replicate. \\

\texttt{RadEval} brings the following solutions:
\begin{itemize}
  \item \textbf{Unified open-source codebase}: every factuality-oriented metric proposed to date is re-implemented in a single, reproducible repository.
  \item \textbf{Metric refinements}: corrected and improved versions of existing metrics offer more faithful estimates of clinical correctness (Section~\ref{radevalbertscore} and Appendix~\ref{sec:green}).
  \item \textbf{Expanded expert test set}: an updated, radiologist-annotated corpus enabling fine-grained studies of how automatic metrics align with human judgments (Section~\ref{sec:test_set}).
  \item \textbf{Ready-made baselines}: published predictions from several widely-cited models are included so new systems can be benchmarked out-of-the-box.
  \item \textbf{Built-in statistical testing}: permutation and bootstrap tests (mirroring best practices in image captioning) enable users to determine whether score differences are statistically significant.
\end{itemize}

\section{Existing Radiology Report Metrics}\label{sec:metrics}
\subsection{Lexical Overlap Metrics}
Early evaluation methods focused on string-level overlap between generated and reference reports, typically using metrics from the natural language processing (NLP) literature. ROUGE~\cite{lin2004rouge} and BLEU~\cite{papineni-etal-2002-bleu} remain among the most common, measuring n-gram precision and recall. METEOR~\cite{banerjee-lavie-2005-meteor} and CIDEr~\cite{vedantam2015cider} have also been adapted from image captioning literature. These metrics are straightforward to compute and require no domain-specific annotation or models, making them popular baselines.

In addition, BERTScore~\cite{zhang2019bertscore} has been proposed to address limitations of n-gram overlap metrics. Instead of relying on exact token matches, it computes semantic similarity between candidate and reference reports using contextualized embeddings from a pretrained language model (\textit{e.g.}, BERT~\cite{devlin2019bert}). Token-level similarity is calculated based on the sum of cosine similarities, offering improved sensitivity to semantic alignment and lexical variation.

However, such metrics perform poorly in RRG settings due to their insensitivity to paraphrasing, semantic equivalence, or clinical correctness. Radiology reports are often sparse, redundant, or variable in linguistic expression, which causes n-gram metrics to underestimate report quality even when the clinical meaning is preserved.

\subsection{Clinical Concept-Based Metrics}
To improve domain specificity, several works introduced evaluation metrics based on clinical concept extraction and comparison. F1CheXbert~\cite{smit2020combining} utilizes a rule-based labeler that extracts 14 predefined CheXpert disease categories~\cite{irvin2019chexpertlargechestradiograph} from both reference and generated reports, then computes an F1 score over presence/absence of these labels. SRR-BERT~\cite{delbrouck2025automatedstructuredradiologyreport} expands the label space to 55 labels, and also supports evaluation on structured reports.

F1RadGraph~\cite{delbrouck-etal-2022-improving} also offers a more expressive alternative by representing reports as structured graphs of anatomical and observational entities and relations. A pretrained graph extraction model is applied to both candidate and reference reports, and graph-level overlap metrics (\textit{e.g.}, precision, recall, F1) are computed. Although F1RadGraph has improved generality and some alignment with radiologist evaluations, it remains limited by the accuracy of the underlying parser and the quality of the training data. While the initial version of F1RadGraph was proposed by \citet{jain2021radgraphextractingclinicalentities}, which computed entity and relation overlap separately and reported their average, it did not consider whether entities were matched based on textual spans, semantic types, or shared relations. This simplification may lead to overestimated alignment in complex cases.

RaTEScore~\cite{zhao-etal-2024-ratescore} identifies medical entities and their types (\textit{e.g.}, anatomy, disease), applies synonym-aware semantic matching, and weights entities by diagnostic importance to better handle terminology variation and negation.

Similar to BERTScore, the CheXbert vector similarity metric~\cite{Yu2022.08.30.22279318} measures alignment between generated and reference reports by computing the cosine similarity of their embeddings obtained via the CheXbert model~\cite{smit-etal-2020-combining}.

In contrast to these standard clinical metrics, Temporal Entity F1~\cite{zhang-etal-2025-libra} is designed to assess temporal information quality in reports. It focuses on capturing the progression or stability of observations (\textit{e.g.}, worsening, improved, or stable)~\cite{bannur2023learningexploittemporalstructure}, and is particularly useful for detecting temporal hallucinations.

\subsection{Composite and Learned Metrics}
To improve alignment with human evaluations, ensemble-based or regression-trained metrics such as RadCliQ~\cite{yu2023evaluating} have been proposed. RadCliQ uses a linear model trained on radiologist-labeled error counts, combining submetrics like BLEU, BERTScore, CheXbert vector similarity, and F1RadGraph. This strategy improves correlation with expert assessments but introduces interpretability challenges.

\subsection{Generative and LLM-Based Evaluation Metrics}
More recently, LLM-based generative evaluation has emerged as a promising paradigm for RRG metric design. These approaches leverage foundation models' reasoning and cross-domain generalization to assess report correctness, style, and completeness in a free-text or structured manner.

These LLM-driven evaluators offer high flexibility and often exhibit better alignment with radiologist judgment, especially on nuanced and out-of-distribution findings. For instance, GREEN~\cite{ostmeier-etal-2024-green} proposed an interpretable and open-source LLM-based evaluation pipeline using a ~7B parameter models, and includes a normalized GREEN score, structured error summaries for interpretability, and zero-shot generalization across imaging modalities.

CheXprompt~\cite{zambrano2025clinically} is a GPT-based evaluator that detects and categorizes six types of clinically relevant errors: false positive and false negative findings, incorrect location or severity, false positive comparisons, and false negative comparisons.

Similarly, FineRadScore~\cite{huang2024fineradscore} evaluates reports line by line, combining clinical severity with the number of incorrect lines. This reflects both the potential clinical risk and the effort required for correction, providing a practical measure of report quality.

RadFact~\cite{bannur2024maira} is also a GPT-based evaluation suite that assesses the factuality of each sentence in a generated report based on the corresponding reference sentences. It supports grounded evaluation and provides interpretable, sentence-level error analysis.

Despite their flexibility and strong alignment with expert judgment, GPT-based evaluation methods face key limitations: high computational cost, deployment barriers due to model size or proprietary APIs, and potential inconsistencies in output. Clinical applications also raise data privacy concerns. RadEval addresses these challenges by standardizing interfaces and supporting lightweight, open-source alternatives like GREEN, enabling local, privacy-preserving, and reproducible evaluation for radiology report generation.

\section{RadEval}\label{sec:radeval}

One of the critical obstacles to building AI systems that can match the accuracy and nuance of expert radiologists is the lack of standardized evaluation metrics. This gap hinders reliable analysis and comparison across different studies, and limits the real-world applicability of research progress.

Despite rapid innovation in metric development, practical barriers remain for adoption and comparison:
\begin{itemize}
    \item Each metric typically requires separate installation, dependencies, and data pre-processing pipelines.
    \item Some tools lack public code or require proprietary APIs.
    \item Evaluation outputs vary in format and interpretability.
\end{itemize}

To mitigate this fragmentation, we propose a system that consolidates access to a wide range of RRG metrics, spanning from n-gram baselines to modern LLM evaluators. This system is designed to be modular, supporting plug-and-play integration of new metrics. By democratizing access to high-quality RRG evaluation tools, we aim to accelerate research on radiology report generation and encourage more standardized and reproducible benchmarking.

\section{Benchmarking}\label{sec:benchmarking}
We conducted extensive benchmarking and evaluation of various models on publicly available datasets (Table~\ref{tab:scores}).

\subsection{Datasets}\label{sec:dataset}
For evaluation, we utilized the official test splits of MIMIC-CXR~\cite{johnson2019mimic} and ReXGradient-160K~\cite{zhang2025rexgradient160klargescalepubliclyavailable}, as well as the public validation set of CheXpert Plus~\cite{chambon2024chexpertplusaugmentinglarge}, as no official test split is available for the latter.

Each study in these datasets may include multiple associated images, all of which were retained for evaluation. Depending on model support, either a single representative image or all available images were used as input. We focused on specifically evaluating the generation of the ``Findings'' and ``Impression'' sections. Reports missing either section were excluded to ensure consistent evaluation across metrics.

\paragraph{MIMIC-CXR}\quad 
A widely used public dataset containing 377,110 chest X-ray images across 227,835 studies, each paired with a radiology report~\cite{johnson2019mimiccxrjpglargepubliclyavailable}. We use JPEG images from MIMIC-CXR-JPG instead of the original DICOMs. The official test split includes 2,347 studies with ``Findings'' sections and 2,224 with ``Impression'' sections.

\paragraph{CheXpert-Plus}\quad 
Comprises 223,462 image–report pairs from 187,711 studies across 64,725 patients~\cite{chambon2024chexpertplusaugmentinglarge}. We use its validation set, which contains 74 studies with ``Findings'' and 234 with ``Impression'' sections.

\paragraph{ReXGradient-160K}\quad 
The largest publicly available chest X-ray dataset to date in terms of patient coverage, including 160,000 image–report pairs from 109,487 patients across 79 U.S. medical sites~\cite{zhang2025rexgradient160klargescalepubliclyavailable}. Its official test set includes 10,000 studies with both ``Findings'' and ``Impression'' sections.

\subsection{Baselines}\label{sec:baselines}
To evaluate the performance of existing RRG systems, we include a representative set of baselines from different institutions and architectures:

\paragraph{CheXpert-Plus}~\citep{chambon2024chexpertplusaugmentinglarge}\quad 
Uses a Swinv2-based vision encoder and a two-layer BERT decoder. Two separate models are trained on MIMIC-CXR: one for the \textit{Findings} section and another for the \textit{Impression}.

\paragraph{CheXagent}~\citep{chen2024visionlanguagefoundationmodelenhance}\quad 
An instruction-tuned foundation model for chest X-ray interpretation. It integrates a vision encoder with a cross-modal adapter to align visual and textual information. Training is conducted on CheXinstruct~\cite{chen2024visionlanguagefoundationmodelenhance}, a diverse instruction dataset aggregated from 28 open-source medical datasets.

\paragraph{MAIRA-2}~\cite{bannur2024maira2groundedradiologyreport}\quad
A model designed for grounded radiology report generation, which involves not only producing clinically accurate text but also identifying the spatial locations of findings. It builds on the LLaVA architecture~\cite{liu2023visualinstructiontuning}, incorporating a frozen Rad-DINO-MAIRA-2 vision encoder~\cite{P_rez_Garc_a_2025}, a Vicuna-7B~\cite{zheng2023judgingllmasajudgemtbenchchatbot} language model, and a four-layer fully connected multilayer perceptron for vision-language alignment.

\paragraph{Libra}~\citep{zhang-etal-2025-libra}\quad
A temporally-aware multimodal large language model (MLLM) tailored for generating the \textit{Findings} section in radiology reports. Unlike prior single-image methods, Libra leverages paired chest X-rays to capture disease progression. It integrates a frozen Rad-DINO~\cite{pérezgarcía2024raddinoexploringscalablemedical} image encoder with the Meditron-7B~\cite{chen2023meditron70bscalingmedicalpretraining} language model via a Temporal Alignment Connector, which combines a Layerwise Feature Extractor and a Temporal Fusion Module to embed multi-scale visual changes over time into the model architecture.

\paragraph{Med-CXRGen}~\citep{zhang-etal-2024-gla}\quad 
Built on LLaVA-Med~\cite{li2023llavamedtraininglargelanguageandvision}, this model uses multi-stage visual instruction tuning and stitches multiple images for unified encoding. Separate models are trained for the ``Findings'' and ``Impression'' sections using the RRG24 dataset~\cite{xu-etal-2024-overview}.

\section{\texttt{RadEval}BERTScore}\label{radevalbertscore}

\begin{table*}[t]
\centering
\small
\setlength\tabcolsep{3pt}
\renewcommand\arraystretch{1.2}
\resizebox{\textwidth}{!}{%
\begin{tabular}{l*{10}{c}}
\toprule
\multirow{2.5}{*}{\textbf{Models}} &
\multicolumn{5}{c}{\textbf{CheXpert 5$\times$200}} &
\multicolumn{5}{c}{\textbf{HOPPR 8$\times$200}} \\
\cmidrule(lr){2-6}\cmidrule(lr){7-11}
& P@5 & P@10 & P@50 & P@100 & mAP &
P@5 & P@10 & P@50 & P@100 & mAP \\
\midrule
\citet{devlin2019bert}            & 46.6 ± 5.5 & 44.8 ± 4.8 & 37.0 ± 2.8 & 33.4 ± 2.1 & 30.1 ± 1.6 &
                              28.1 ± 2.1 & 24.6 ± 2.0 & 19.1 ± 1.0 & 17.3 ± 0.8 & 15.6 ± 0.4 \\
\citet{warner2024smarter}             & 39.4 ± 4.1 & 35.7 ± 4.1 & 30.5 ± 2.2 & 28.3 ± 1.7 & 26.5 ± 1.2 &
                              23.0 ± 2.1 & 20.3 ± 1.6 & 16.8 ± 0.5 & 15.8 ± 0.5 & 14.6 ± 0.2 \\
\citet{sounack2025bioclinical}  & 38.6 ± 4.6 & 36.6 ± 3.9 & 30.6 ± 2.3 & 28.2 ± 1.6 & 25.6 ± 0.9 &
                              23.4 ± 2.3 & 21.4 ± 1.6 & 17.4 ± 1.0 & 15.9 ± 0.5 & 14.7 ± 0.3 \\
\texttt{RadEval}BERT                       & 60.3 ± 3.1 & 56.4 ± 2.6 & 46.4 ± 2.0 & 41.4 ± 1.7 & 36.4 ± 1.2 &
                              38.6 ± 2.0 & 34.8 ± 2.2 & 26.9 ± 1.0 & 23.7 ± 1.0 & 20.1 ± 0.6 \\
\bottomrule
\multicolumn{8}{c}{\textbf{CheXbert Test Set}}  \\
\cmidrule(lr){2-6}
\multirow{-2}{*}{\textbf{Models}} & P@5 & P@10 & nDCG@5 & nDCG@10 & mAP \\
\cmidrule(r{0em}){0-5}
\citet{devlin2019bert}           & 64.3 ± 2.0 & 59.2 ± 1.8 & 54.5 ± 1.3 & 48.7 ± 1.0 & 50.1 ± 1.1 \\
\citet{warner2024smarter}             & 62.0 ± 2.1 & 57.2 ± 1.3 & 51.8 ± 1.5 & 46.2 ± 0.9 & 48.7 ± 1.0 \\
\citet{sounack2025bioclinical}  & 61.8 ± 1.0 & 56.8 ± 1.0 & 51.2 ± 1.0 & 45.8 ± 0.8 & 48.3 ± 0.9 \\
\texttt{RadEval}BERT                      & 70.2 ± 2.2 & 64.6 ± 1.5 & 58.9 ± 1.6 & 53.3 ± 1.1 & 53.4 ± 1.0 \\
\end{tabular}}
\caption{Zero-shot report-to-report retrieval performance of \texttt{RadEval}BERTScore Evaluation. We evaluate on three datasets: CheXpert 5×200 (five single-label categories), HOPPR 8×200 (eight out-of-domain single-label categories), and the CheXbert multi-label test set. For CheXpert and HOPPR, we report Precision@$\{5,10,50,100\}$ and mean Average Precision (mAP); for CheXbert we report Precision@$\{5,10\}$, normalized Discounted Cumulative Gain (nDCG@$\{5,10\}$), and mAP. Values are presented as mean ± standard deviation over 10 random seeds.}
\label{tab:retrieval-prec-perc}
\end{table*}

In this work, we also introduce a domain-specific radiology language encoder trained using SimCSE ~\cite{gao2022simcsesimplecontrastivelearning} -- a contrastive learning method for sentence embeddings -- to capture high-quality representations of radiology report text. We begin with a \texttt{ModernBERT-base} architecture and train it on the ``Findings'' and ``Impression'' sections from MIMIC-CXR, CheXpert, and ReXGradient-160K. This pretraining setup allows the model to learn clinically meaningful semantic relationships within and across radiology reports. 

We demonstrate the effectiveness of our embedding model on a
\textbf{zero-shot report-to-report retrieval} task.
The set-up mirrors a classic text-retrieval scenario:

\begin{enumerate}
  \item A small set of \textbf{query reports} and a larger pool of
        \textbf{candidate reports}, each annotated with one or more radiology
        labels, are encoded with a frozen text encoder.
  \item For every query, we compute the cosine similarity to \emph{all}
        candidates and obtain a ranked list in descending similarity order.
\end{enumerate}

A candidate is considered {relevant} to a query
if and only if the two reports share {at least one} label. For each of our experiments, we choose 10 queries and up to 200 positive candidates. 

\subsection{Metrics}

\begin{description}
  \item[Precision@\(K\)]\hfill\\
        Fraction of the first \(K\) retrieved reports that are relevant.
        A hit is counted as soon as \emph{one} label overlaps with the query.

  \item[mean Average Precision (mAP)]\hfill\\
        For each query, we compute \emph{Average Precision} (\textit{i.e.}, the mean of the precision values at every rank where a relevant report occurs. The scan stops at the {last} relevant rank, so items appearing afterwards cannot influence the score. mAP is the mean AP over all queries and reflects {how early, on average, relevant reports are surfaced}.

  \item[nDCG@\(K\)]\hfill\\
        A graded variant that rewards richer matches.
        The gain between a query and a candidate is the
        {number of shared labels}.
        DCG is accumulated over the top \(K\) positions with a logarithmic
        discount \(1/\log_2(\text{rank}+1)\);
        nDCG normalises by the ideal DCG, yielding a score in \([0,1]\)
        where 1 means the system ranks the strongest overlaps highest.
\end{description}

\subsection{Datasets}

\begin{description}
  \item[CheXpert~5$\times$200]\hfill\\
        Five single-label categories  
        (Atelectasis, Cardiomegaly, Edema, Consolidation, Pleural~Effusion). \\
        $\rightarrow$We report Precision@$\{5,10,50,100\}$ and mAP.

  \item[CheXbert Test Set]\hfill\\
        Multi-label reports with 14 chexpert possible findings  
        (\textit{e.g.}, Airspace Opacity, Pneumonia, Support Devices). \\
        $\rightarrow$Because at most 20 positives exist per query, we report Precision@$\{5,10\}$ and nDCG@$\{5,10\}$ together with mAP.

  \item[HOPPR~8$\times$200]\hfill\\
        This is an out-of-domain, single-label dataset used to evaluate generalization performance. The label categories include: acute rib fracture, air space opacity, cardiomegaly, lung nodule or mass, non acute rib fracture, pleural fluid, pneumothorax, and pulmonary artery enlargement. \\
        $\rightarrow$We report Precision@$\{5,10,50,100\}$ and mAP.
\end{description}

This protocol provides complementary views of retrieval quality:
Precision@\(K\) for top-\(K\) exact-hit rate,
mAP for overall early-ranking performance,
and nDCG@\(K\) for sensitivity to \emph{how many} labels overlap.

\begin{table*}[p] 
\centering
\small
\renewcommand{\arraystretch}{1.0}
\caption{Top-3 metrics by Kendall’s $\tau_b$ (more negative is better; higher metric $\Rightarrow$ fewer errors).
“\cmark\ aligned” = 95\% CI $<0$; “\xmark\ misaligned” = 95\% CI $>0$; “ns” = CI overlaps 0.
\textbf{Scope:} pooled rows show \texttt{(pairs, n)}; blocked rows show \texttt{(blocks, pairs)}.
Each study has $K{=}3$ candidates (Findings: 148 studies; Impression: 60).}
\label{tab:rexval-top3-compact}
\begin{tabular}{@{}l l r@{\,}l c l@{}}
\toprule
Endpoint & Metric & \multicolumn{2}{c}{$\tau_b$ [95\% CI]} & Sig & Scope \\
\midrule
\multicolumn{6}{@{}l}{\textbf{Overall (pooled) vs.\ total significant errors}}\\
\addlinespace[2pt]
 & green            & $-0.183$ & $[-0.246,-0.122]$ & \cmark\ aligned & (pairs 194{,}376, n 624) \\
 & srr\_bert        & $-0.133$ & $[-0.193,-0.071]$ & \cmark\ aligned & (pairs 194{,}376, n 624) \\
 & radcliq          & $-0.107$ & $[-0.160,-0.052]$ & \cmark\ aligned & (pairs 194{,}376, n 624) \\
 & radevalbertscore & $-0.076$ & $[-0.131,-0.017]$ & \cmark\ aligned & (pairs 194{,}376, n 624) \\
 & rouge            & $-0.038$ & $[-0.091,\,0.017]$ & ns              & (pairs 194{,}376, n 624) \\
 & bertscore        & $-0.027$ & $[-0.085,\,0.032]$ & ns              & (pairs 194{,}376, n 624) \\
 & radgraph         & $-0.011$ & $[-0.068,\,0.048]$ & ns              & (pairs 194{,}376, n 624) \\
 & bleu             & $ 0.034$ & $[-0.020,\,0.086]$ & ns              & (pairs 194{,}376, n 624) \\
 & chexbert         & $ 0.074$ & $[\,0.012,\,0.137]$ & \xmark\ misaligned & (pairs 194{,}376, n 624) \\
\addlinespace[4pt]
\multicolumn{6}{@{}l}{\textbf{ALL (blocked) vs.\ total significant errors}}\\
\addlinespace[2pt]
 & green            & $-0.195$ & $[-0.295,-0.087]$ & \cmark\ aligned & (blocks 159, pairs 477) \\
 & bertscore        & $-0.160$ & $[-0.260,-0.059]$ & \cmark\ aligned & (blocks 188, pairs 564) \\
 & bleu             & $-0.153$ & $[-0.273,-0.029]$ & \cmark\ aligned & (blocks  89, pairs 267) \\
\addlinespace[4pt]
\multicolumn{6}{@{}l}{\textbf{ALL (blocked) vs.\ total insignificant errors}}\\
\addlinespace[2pt]
 & radcliq          & $-0.039$ & $[-0.147,\,0.076]$ & ns & (blocks 143, pairs 429) \\
 & radevalbertscore & $-0.009$ & $[-0.126,\,0.103]$ & ns & (blocks 133, pairs 399) \\
 & bertscore        & $-0.008$ & $[-0.107,\,0.094]$ & ns & (blocks 143, pairs 429) \\
\addlinespace[4pt]
\multicolumn{6}{@{}l}{\textbf{Impression only (blocked) vs.\ total significant errors}}\\
\addlinespace[2pt]
 & bertscore        & $-0.225$ & $[-0.399,-0.049]$ & \cmark\ aligned & (blocks  57, pairs 171) \\
 & rouge            & $-0.215$ & $[-0.383,-0.042]$ & \cmark\ aligned & (blocks  57, pairs 171) \\
 & radevalbertscore & $-0.206$ & $[-0.399,-0.010]$ & \cmark\ aligned & (blocks  53, pairs 159) \\
\addlinespace[4pt]
\multicolumn{6}{@{}l}{\textbf{Findings only (blocked) vs.\ total significant errors}}\\
\addlinespace[2pt]
 & green            & $-0.196$ & $[-0.314,-0.075]$ & \cmark\ aligned & (blocks 113, pairs 339) \\
 & bleu             & $-0.168$ & $[-0.313,-0.020]$ & \cmark\ aligned & (blocks  68, pairs 204) \\
 & bertscore        & $-0.132$ & $[-0.246,-0.017]$ & \cmark\ aligned & (blocks 131, pairs 393) \\
\addlinespace[6pt]
\multicolumn{6}{@{}l}{\textbf{Per-category (blocked, significant-error endpoints)}}\\
\addlinespace[2pt]
\shortstack[l]{significant: false prediction\\ of finding}
 & green            & $-0.082$ & $[-0.198,\,0.030]$ & ns & (blocks 134, pairs 400) \\
 & radcliq          & $-0.052$ & $[-0.157,\,0.052]$ & ns & (blocks 162, pairs 482) \\
 & bertscore        & $-0.049$ & $[-0.158,\,0.061]$ & ns & (blocks 162, pairs 482) \\
\addlinespace[2pt]
\shortstack[l]{significant: omission\\ of finding}
 & srr\_bert        & $-0.512$ & $[-0.625,-0.390]$ & \cmark\ aligned & (blocks  91, pairs 271) \\
 & chexbert         & $-0.503$ & $[-0.626,-0.366]$ & \cmark\ aligned & (blocks  89, pairs 267) \\
 & green            & $-0.250$ & $[-0.374,-0.126]$ & \cmark\ aligned & (blocks 113, pairs 337) \\
\addlinespace[2pt]
\shortstack[l]{significant: incorrect location/\\ position of finding}
 & bertscore        & $-0.068$ & $[-0.213,\,0.079]$ & ns & (blocks  80, pairs 238) \\
 & radevalbertscore & $-0.040$ & $[-0.201,\,0.123]$ & ns & (blocks  76, pairs 226) \\
 & rouge            & $-0.018$ & $[-0.174,\,0.139]$ & ns & (blocks  80, pairs 238) \\
\addlinespace[2pt]
\shortstack[l]{significant: incorrect severity\\ of finding}
 & radevalbertscore & $-0.033$ & $[-0.223,\,0.160]$ & ns & (blocks  56, pairs 168) \\
 & bleu             & $-0.001$ & $[-0.235,\,0.219]$ & ns & (blocks  35, pairs 105) \\
 & rouge            & $ 0.007$ & $[-0.170,\,0.191]$ & ns & (blocks  61, pairs 181) \\
\addlinespace[2pt]
\shortstack[l]{significant: spurious comparison\\ (not in reference)}
 & bertscore        & $-0.153$ & $[-0.300,\,0.001]$ & ns & (blocks  81, pairs 241) \\
 & radcliq          & $-0.125$ & $[-0.263,\,0.014]$ & ns & (blocks  81, pairs 241) \\
 & radevalbertscore & $-0.103$ & $[-0.247,\,0.063]$ & ns & (blocks  77, pairs 229) \\
\addlinespace[2pt]
\shortstack[l]{significant: omission of change\\ from previous study}
 & bleu             & $-0.127$ & $[-0.335,\,0.099]$ & ns & (blocks  37, pairs 111) \\
 & green            & $-0.066$ & $[-0.241,\,0.113]$ & ns & (blocks  65, pairs 195) \\
 & radgraph         & $-0.027$ & $[-0.199,\,0.137]$ & ns & (blocks  61, pairs 183) \\
\addlinespace[2pt]
\shortstack[l]{significant: inarticulate report\\ (grammar/readability)}
 & radcliq          & $-0.350$ & $[-0.560,-0.140]$ & \cmark\ aligned & (blocks  35, pairs 105) \\
 & radevalbertscore & $-0.266$ & $[-0.476,-0.046]$ & \cmark\ aligned & (blocks  34, pairs 102) \\
 & bertscore        & $-0.251$ & $[-0.480,-0.013]$ & \cmark\ aligned & (blocks  35, pairs 105) \\
\bottomrule
\end{tabular}
\end{table*}

\section{\texttt{RadEval} Expert Dataset}\label{sec:test_set}

\paragraph{Dataset.}
We release an updated \texttt{RadEval}-expert dataset with board-certified radiologists annotating clinically significant and insignificant errors across different error categories. Building on ReXVal~\cite{yu2023radiologyrexval}, we annotate false predictions of findings, omissions of findings, incorrect locations/positions, incorrect severities, spurious comparisons, omissions of changes from prior studies, as well as a new category: \textbf{inarticulate report/grammar}. All error categories are labeled as either significant or insignificant. We also extend beyond the ``Impression'' to also cover the ``Findings''' section. The corpus comprises \textbf{208 studies} (\textbf{148 findings} and \textbf{60 impressions}), and \textbf{each study has exactly $K{=}3$ annotated candidate reports per ground truth}. Ground-truth reports come from MIMIC-CXR, CheXpert-Plus, and ReXGradient-160K, and candidate reports are generated by CheXagent, the CheXpert-Plus model, and MAIRA-2.

\paragraph{Methods.}
We measure agreement between automatic metrics and radiologists’ judgments using Kendall’s $\tau_b$~\cite{kendall1945ties} against radiologist error counts.
We report: (1) a \textbf{pooled} (overall) $\tau_b$ versus the \textbf{total number of significant errors} (treating each candidate independently; ignores study grouping), and (2) a \textbf{blocked} (within-study, tie-aware) $\tau_b$ with 95\% {block-bootstrap} confidence intervals obtained by resampling {study} blocks (preserving section type).
Blocked analyses cover: totals of {significant} and {insignificant} errors across all sections (“ALL”), the significant-error total within each section ({Impression}, {Findings}), and each individual {significant-error category}.
We label \cmark\ {aligned} when the 95\% CI lies entirely below $0$ (higher metric $\Rightarrow$ fewer errors), \xmark\ {misaligned} when the CI lies entirely above $0$ (higher scores $\Rightarrow$ more errors), and \textbf{ns} otherwise.

\paragraph{Results.}
Overall (pooled vs.\ total significant errors), \textit{green}, \textit{srr\_bert}, \textit{radcliq}, and \textit{radevalbertscore} show meaningful negative correlations (\cmark aligned), while \textit{rouge}, \textit{bertscore}, \textit{radgraph}, and \textit{bleu} are not significant; \textit{chexbert} is \xmark misaligned (higher scores with more errors).
Within studies (blocked, ALL), \textit{green}, \textit{bertscore}, and \textit{bleu} are top for total significant errors (all \cmark aligned), and no metric tracks total \emph{insignificant} errors (all ns).

Table~\ref{tab:rexval-top3-compact} reports correlations by section and across different error category types. From these results, \textit{green} emerges as the most reliable single metric for tracking clinically significant errors, followed by \textit{srr\_bert}.

\section{Conclusion}\label{sec:conclusion}
We introduced RadEval, a unified framework for evaluating RRG. By consolidating and standardizing a diverse suite of evaluation metrics, including lexical overlap, clinical concept extraction, structured graph comparison, and LLM-based scoring, RadEval addresses longstanding reproducibility and benchmarking challenges in the RRG domain. We refined existing metrics, released a high-fidelity expert-annotated test set, and benchmarked state-of-the-art report generation systems across multiple publicly available datasets. In addition, we demonstrated the utility of a new domain-specific sentence encoder through a zero-shot retrieval task and introduced an updated lightweight version of the GREEN metric capable of cross-modality evaluation. RadEval's modular architecture will help facilitate robust, reproducible, and clinically grounded evaluation -- ultimately helping accelerate the safe deployment of radiology AI systems.

\section*{Limitations}\label{sec:limitations}
While RadEval already unifies a broad set of automated radiology text evaluation metrics, recently proposed LLM-based metrics (\textit{e.g.}, CheXprompt, FineRadScore, and RadFact) are not currently implemented due to their reliance on LLM APIs or lack of standardization -- though they remain valuable future additions. Additionally, some metrics depend on upstream parsers that may introduce noise. Current evaluations also focus primarily on chest X-ray radiology reports written in English from institutions in the U.S., limiting generalizability across languages and geographical regions. Finally, we aim to continue to expand the expert-labeled dataset with additional reports and clinical annotators, as well as to compute detailed correlation analyses across all automated metrics and annotations to better assess metric alignment with clinical judgment.


\bibliography{custom}
\appendix
\clearpage
\section{Score table}

\begin{strip}
\centering
\captionof{table}{Benchmarking results across multiple models on standard datasets under default system prompts.}
\label{tab:scores}
\begin{adjustbox}{angle=90,max width=\textwidth,max height=.91\textheight}
  \begin{minipage}{\textheight}
    \begin{tabularx}{\textheight}{l|YYYYYYYYYYYYYY}
      \toprule
      \textbf{Models}
        & \makecell[t]{GREEN\\Llama}
        & \makecell[t]{ROUGE\\L}
        & \makecell[t]{\\F1\\RG}
        & \makecell[t]{\\RG\\ER}
        & \makecell[t]{\\RG\\$\overline{ER}$}
        & \makecell[t]{\\\\BLEU}
        & \makecell[t]{\\BERT\\Score}
        & \makecell[t]{\\SRR\\BERT}
        & \makecell[t]{\\F1cXb\\5}
        & \makecell[t]{\\F1cXb\\14}
        & \makecell[t]{\\RaTE\\Score}
        & \makecell[t]{Rad\\CLiQ\\v1}
        & \makecell[t]{Temp.\\Entity\\F1}
        & \makecell[t]{Rad\\Eval\\BERT} \\
        \hline
        \multicolumn{15}{c}{\textbf{CheXpert-Plus Validation Findings}} \\
        \hline
        CheXagent          & 26.2 & 21.4 & 27.4 & 24.8 & 17.6 &  4.1 & 33.9 & 50.5 & 55.8 & 58.1 & 51.1 & 67.9 & 40.0 &  8.2 \\
        CheXpert-Plus      & 25.2 & 21.7 & 25.8 & 22.9 & 17.2 &  4.7 & 47.1 & 46.9 & 48.7 & 52.8 & 51.0 & 77.3 & 34.4 & 12.7 \\
        Med-CXRGen-F  & 25.5 & 22.1 & 23.7 & 21.2 & 15.5 &  6.4 & 48.0 & 44.5 & 41.2 & 47.6 & 50.1 & 75.3 & 32.6 & 14.2 \\
        Libra-v1.0-3B        & 21.9 & 18.8 & 18.6 & 15.9 & 11.0 &  2.8 & 42.9 & 47.1 & 50.9 & 50.7 & 47.0 & 65.1 & 19.1 & 10.9 \\
        Libra-v1.0-7B        & 22.0 & 18.0 & 17.9 & 16.2 & 11.0 &  2.5 & 43.8 & 44.1 &    47.8 &    49.1 & 47.3 &    65.6 & 26.0 & 11.5 \\
        MAIRA-2             & 15.5 & 16.4 & 14.6 & 13.3 &  9.2 &  1.9 & 41.8 & 41.0 &    44.0 &    47.1 & 44.0 &    63.6 & 29.1 & 10.7 \\
        \hline
        \multicolumn{15}{c}{\textbf{CheXpert-Plus Validation Impressions}} \\
        \hline
        CheXpert-Plus      & 23.7 & 25.3 &  3.4 &  3.1 &  8.0 &  2.9 & 51.2 & 44.5 & 45.7 & 49.0 & 47.7 & 48.8 & 35.6 & 14.8 \\
        Med-CXRGen-I  & 26.6 & 23.6 & 18.7 & 16.7 & 12.6 &  6.5 & 50.8 & 44.4 & 48.2 & 48.3 & 45.5 & 64.6 & 35.3 & 20.0 \\
        \hline
        \multicolumn{15}{c}{\textbf{MIMIC-CXR Test Findings}} \\
        \hline
        CheXagent          & 29.6 & 23.1 & 26.8 & 24.2 & 17.4 &  4.9 & 39.0 & 49.7 & 60.1 & 55.1 & 56.0 & 71.6 & 22.5 & 37.2 \\
        CheXpert-Plus      & 30.6 & 25.7 & 24.2 & 22.1 & 17.0 &  5.7 & 54.2 & 48.2 & 54.1 & 47.4 & 54.2 & 84.6 & 22.5 & 46.8 \\
        Med-CXRGen-F  & 28.1 & 22.7 & 20.9 & 18.6 & 13.8 &  5.9 & 50.4 & 44.6 & 53.3 & 45.2 & 52.4 & 75.0 & 16.8 & 44.1 \\
        Libra-v1.0-3B        & 29.2 & 21.7 & 20.3 & 18.0 & 12.9 &  5.1 & 49.4 & 46.4 & 60.0 & 52.5 & 53.5 & 76.0 & 14.5 & 46.4 \\
        Libra-v1.0-7B        & 28.0 & 20.9 & 19.9 & 17.5 & 12.3 &  4.6 & 49.5 & 45.8 & 61.5 & 53.7 & 52.7 & 75.1 & 18.0 & 46.5 \\
        MAIRA-2             & 22.4 & 18.5 & 17.3 & 15.3 & 10.9 &  3.0 & 47.6 & 42.8 & 59.1 & 50.7 & 51.0 & 71.1 & 16.4 & 40.2 \\
        \hline
        \multicolumn{15}{c}{\textbf{MIMIC-CXR Test Impressions}} \\
        \hline
        CheXpert-Plus      & 25.0 & 23.6 & 22.6 & 20.1 & 16.8 &  2.9 & 46.2 & 36.8 & 50.4 & 45.3 & 46.4 & 84.6 & 37.7 & 29.9 \\
        Med-CXRGen-I & 20.8 & 14.6 & 13.5 & 11.7 &  8.5 &  2.3 & 39.3 & 35.4 & 49.1 & 41.5 & 43.4 & 61.4 & 16.7 &  9.7 \\
        \hline
        \multicolumn{15}{c}{\textbf{ReXGradient-160K Test Findings}} \\
        \hline
        CheXagent          &    44.5 & 21.7 & 25.9 & 24.5 & 21.0 &  3.3 & 39.2 & 41.0 &  4.5 &  5.5 & 59.2 &    69.7 & 57.1 & 36.4 \\
        CheXpert-Plus      &    33.6 & 23.5 & 22.0 & 20.7 & 16.8 &  3.5 & 50.9 & 45.4 & 16.5 & 21.0 & 55.0 &    76.7 & 45.8 & 36.5 \\
        Med-CXRGen-F  &    29.6 & 14.7 & 10.1 &  9.2 &  4.4 &  1.6 & 42.0 & 39.7 &  0.7 & 13.0 & 43.6 &    65.8 & 62.1 & 33.1 \\
        Libra-v1.0-7B        &    44.8 & 21.9 & 20.5 & 18.5 & 13.1 &  3.8 & 51.0 & 45.7 &  7.4 & 15.7 & 56.2 &    78.7 & 40.5 & 40.6 \\
        MAIRA-2             &    36.5 & 21.4 & 20.1 & 18.3 & 14.3 &  2.3 & 48.9 & 43.0 &  4.1 & 14.0 & 52.9 &    77.9 & 60.8 & 35.5 \\
        \hline
        \multicolumn{15}{c}{\textbf{ReXGradient-160K Test Impressions}} \\
        \hline
        CheXpert-Plus      &    3.8 &  7.6 &  2.5 &  1.9 &  1.2 &  0.2 & 30.3 & 37.8 & 16.3 &  9.4 & 30.1 &    51.2 & 21.2 & 18.7 \\
        Med-CXRGen-I  &    22.3 &  8.2 &  3.9 &  3.6 &  2.4 &  0.4 & 31.5 & 40.6 &  2.6 & 24.4 & 31.4 &    51.7 & 55.6 & 10.5 \\
        \hline
    \end{tabularx}
  \end{minipage}
\end{adjustbox}
\end{strip}

\clearpage

\section{Refined GREEN Metric}\label{sec:green}
To extend the capabilities of the GREEN metric, we finetuned a compact Gemma-2B model on the original GREEN dataset along with an additional ~50,000 annotated radiology report pairs spanning multiple imaging modalities, including CT, MRI, and ultrasound. This represents an evolution beyond the original implementation, which focused exclusively on chest X-rays. By leveraging a smaller, more lightweight language model, we achieve substantial improvements in computational efficiency -- averaging inference times of 2–3 seconds per report -- while maintaining performance comparable to larger models such as Llama and Phi. This reduced resource footprint makes the updated GREEN model more practical for large-scale or real-time deployment settings. Our work underscores the potential for continued improvement by incorporating more diverse imaging contexts and exploring even lighter architectures without sacrificing clinical alignment or interpretability.

\end{document}